\documentclass[letterpaper, 10 pt, conference]{ieeeconf}  

\IEEEoverridecommandlockouts                              

\usepackage{amsmath} 
\usepackage{amssymb}  
\usepackage{bm}
\usepackage{cite}
\usepackage{hyphenat}
\usepackage{graphicx}
\usepackage{booktabs}
\usepackage{subfig}
\usepackage{color}
\usepackage{siunitx}
\usepackage{marvosym}
\usepackage[normalem]{ulem}
\usepackage{multirow}
\usepackage{caption}[small]
\usepackage{dblfloatfix}
\usepackage{hyperref}

\iftrue
\definecolor{orange}{rgb}{0.8,0.5,0.0} 

\else

\fi
\newcommand{\videolink}{https://youtu.be/q4iuobMbjME}

\title{\LARGE \bf
An End-to-End Learning-Based Multi-Sensor Fusion for Autonomous Vehicle Localization}

\author{Changhong Lin, Jiarong Lin$^\dag$, Zhiqiang Sui, XiaoZhi Qu, Rui Wang, Kehua Sheng, Bo Zhang
\thanks{The authors are with the DiDi Autonomous Driving, DiDi Chuxing, Beijing, China.
\{linchanghong, zivlin, suizhiqiang, quxiaozhi, raywangrui, shengkehua, zhangbo\}@didiglobal.com
}
\thanks{$^\dag$Corresponding author: Jiarong Lin}
}

\begin{document}

\maketitle
\thispagestyle{empty}
\pagestyle{empty}

\begin{abstract}

Multi-sensor fusion is essential for autonomous vehicle localization, as it is capable of integrating data from various sources for enhanced accuracy and reliability.
The accuracy of the integrated location and orientation depends on the precision of the uncertainty modeling. Traditional methods of uncertainty modeling typically assume a Gaussian distribution and involve manual heuristic parameter tuning. However, these methods struggle to scale effectively and address long-tail scenarios.
To address these challenges, we propose a learning-based method that encodes sensor information using higher-order neural network features, thereby eliminating the need for uncertainty estimation.
This method significantly eliminates the need for parameter fine-tuning by developing an end-to-end neural network that is specifically designed for multi-sensor fusion. In our experiments, we demonstrate the effectiveness of our approach in real-world autonomous driving scenarios. Results show that the proposed method outperforms existing multi-sensor fusion methods in terms of both accuracy and robustness. A video of the results can be viewed at \url{\videolink}.

\end{abstract}

\section{Introduction}

The Global Navigation Satellite Systems (GNSS) are widely used to provide absolute positioning. However, GNSS has several limitations that impact its effectiveness in autonomous driving, including signal blockage, the multi-path effect, and atmospheric interference \cite{hofmann2007gnss}. To address these limitations, multi-sensor fusion systems that combine GNSS, Inertial Measurement Unit (IMU), and wheel speedometers have emerged as vital solutions. By leveraging the strengths of these different sensors, these systems enhance the reliability and accuracy of localization, ensuring robust performance even in challenging environments.

Existing methods for multi-sensor fusion are predominantly based on Bayesian filters-based such as the Kalman filters \cite{Kalman_filter}, particle filters \cite{Particle_filter}, and graph-based optimization techniques \cite{dellaert2017factor}. The performance of these methods relies heavily on the accurate modeling of sensor measurements, particularly the correctness of the uncertainty associated with each input measurement. Different types of sensor modalities, such as GNSS and IMU, have unique sources of uncertainty that arise from factors related to hardware, environmental conditions, and data processing. Specifically, GNSS measurements encounter issues like signal occlusion in urban areas, while IMUs experience drift over time, especially in dynamic driving conditions. 
However, Bayesian filter-based approaches typically assume a Gaussian distribution and rely on heuristic parameter tuning, which may not accurately capture highly nonlinear real-world scenarios. This leads to challenges in scaling these methods and effectively addressing long-tail scenarios across diverse environments.

\begin{figure}
\centering
\includegraphics[width=1\linewidth]{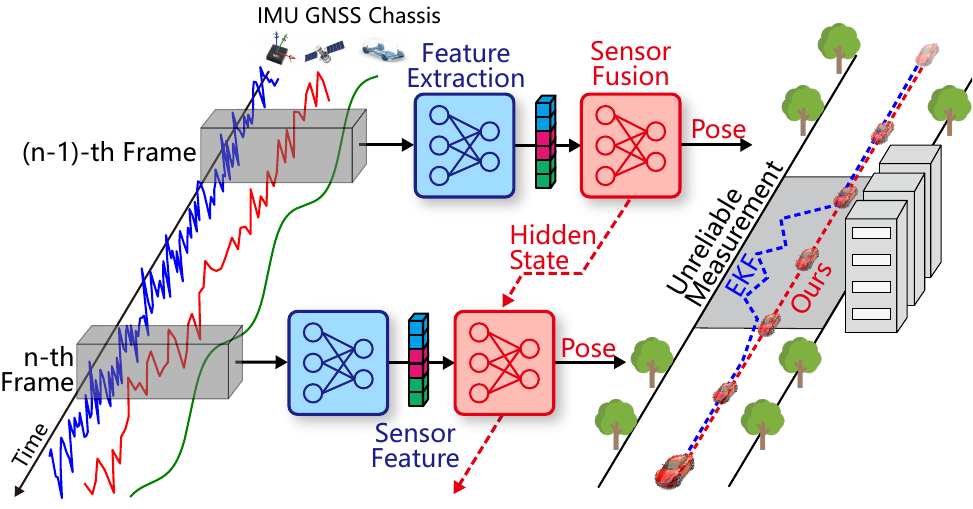}
\caption{An overview of our end-to-end network for multi-sensor fusion. Our experiments demonstrate superior performance in scenarios with unreliable measurements (e.g., GNSS signal blockage), as shown on the right side of the figure.}
\label{fig:architecture}
\vspace{-0.3cm}
\end{figure}

To address these challenges, an end-to-end approach is proposed by leveraging the deep neural network representation and generalization ability for multi-sensor fusion, as depicted in Fig. \ref{fig:architecture}. To the best of our knowledge, this is the first method to use end-to-end learning-based multi-sensor (i.e., GNSS-IMU-Chassis) fusion for localization.
In summary, the key contributions of our work are:
\begin{itemize}
\item An end-to-end multi-sensor fusion framework is proposed, utilizing a Recurrent Neural Network (RNN) to process input sensor measurements and directly output the desired vehicle pose.
\item Sensor information is encoded using neural networks, eliminating the need for sensor noise covariance estimation and avoiding the parameter tuning required by existing model-based methods.
\item The proposed framework is evaluated through real-world experiments, demonstrating its superiority over state-of-the-art methods in both accuracy and robustness.
\end{itemize}

\section{Related Work}
The general solution for multi-sensor fusion follows Bayesian filters, such as particle filters \cite{Particle_filter}, Kalman filters \cite{Kalman_filter}, and extended Kalman filters (EKF)\cite{EKF}, which consist of predict and update steps in terms of Markov chain theory for state transition. The uncertainy, usually presented as process noise and measurement noise, is the key for accuracy and robustness. To estimate the uncertainty, some methods \cite{innovation_based_Adaptive_KF, feng2014kalman} aim to adaptively adjust the process and measurement noise covariance matrices based on innovation information. Besides, data-driven methods have recently been utilized to model measurement covariance across various sensor types, including point clouds \cite{CELLO-3D}, camera images \cite{DICE, Backprop-kf, Multivariate_Uncertainty}, GNSS measurements \cite{NLOS, Deep_RTK}, and pseudo-measurements \cite{AI-IMU}.

In combined GNSS/IMU systems, some methods employ neural networks to process sequential sensor data for estimating covariance matrices, which are then integrated into the EKF. Certain approaches estimate only the process noise covariance \cite{RL-AKF, Fuzzy, A-KIT, TLIO}, while others estimate both process and measurement noise covariance matrices \cite{MT-AKF, Trainable_Quaternion, Du2023Agrobot}.
One of the state-of-the-art methods is \textit{Agrobot} \cite{Du2023Agrobot}, which employs a Kalman filter-based GNSS/INS navigation system tailored for agricultural vehicles. In this system, a Temporal Convolutional Network (TCN) estimates velocity and its noise covariance from subsequences of IMU acceleration, angular velocity, and magnetometer measurements, feeding this information into a Kalman filter to enhance vehicle position and velocity estimation, thereby reducing over-reliance on GNSS.
Leveraging the expressive capabilities of neural networks to capture sensor characteristics, these methods have shown improved performance over traditional EKF approaches. 
However, these methods still rely on the Gaussian distribution assumption to represent uncertainty using covariance matrices. Furthermore, to ensure that the covariance matrix remains positive definite, some approaches assume orthogonality among all variables and estimate only the diagonal terms, limiting the neural network's capability of sensor modeling. Additionally, embedding the EKF into neural network training makes convergence challenging.

Unlike the aforementioned methods, our approach eliminates the tedious task of covariance estimation for sensors by encoding sensor information using neural network features, moving beyond the traditional covariance matrix to achieve better performance. This approach enables an integrated end-to-end localization system that effectively addresses multisensor fusion.

\section{Proposed Method}
In this section, we first present the overview of our proposed algorithm, and then describe the key modules of the algorithms.

\subsection{System Overview}

\begin{figure}
\centering
\includegraphics[width=1\linewidth]{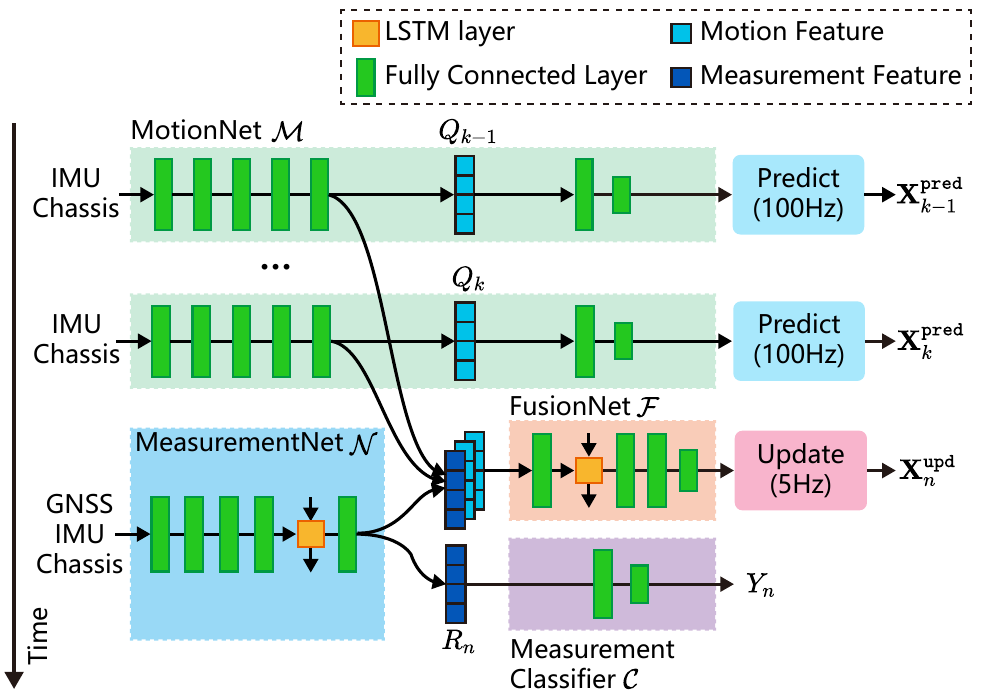}
\caption{The proposed architecture processes and fuse sensor data to output vehicle poses with neural network.}
\vspace{-0.70cm}
\label{fig:network_structure}
\end{figure}

The structure of our proposed framework is depicted in Fig.~\ref{fig:network_structure}, which contains a prediction and an update stage to estimate the vehicle state $\mathbf{X}$, defined as below:
\begin{equation}
\mathbf{X} = [x, y, v, \psi]^T \label{eq:state_def}
\end{equation}
where $x$ and $y$ denote the coordinates in the global coordinate system, $v$ represents the vehicle velocity, where negative values indicate reversing, and $\psi$ indicates the vehicle heading.

In the prediction stage, \textit{MotionNet}, denoted as $\boldsymbol{\mathcal{M}}$, processes control variables and outputs the predicted state and motion feature. When new measurements arrive in the update stage, \textit{MeasurementNet}, denoted as $\boldsymbol{\mathcal{N}}$, assesses the measurement signals to extract their features. Finally, \textit{FusionNet}, denoted as $\boldsymbol{\mathcal{F}}$, combines the motion and measurement features to estimate the fused state. All submodules in our system are constructed with Multi-Layer Perceptron (MLP) network and RNN. The MLP network comprises multiple layers with dropout layers. The RNN network includes an MLP encoder, a unidirectional Long Short-Term Memory (LSTM) component, and, optionally, an MLP decoder.

\subsection{MotionNet}

The \textit{MotionNet} $\boldsymbol{\mathcal{M}}$ can be described as (\ref{eq:motion_net}):

\begin{equation}
\mathbf{F} = \boldsymbol{\mathcal{M}}([\mathbf{I}_{i}, \mathbf{I}_{c}])
\label{eq:motion_net}
\end{equation}

\noindent where $\mathbf{F}=[\delta \mathbf{u}, \mathbf{Q}]^T$ is the model output, which consists control variables residuals $\delta \mathbf{u}$ and the motion features $\mathbf{Q}$. $\mathbf{Q}$ captures the motion feature and will be used in sensor fusion (in Section \ref{sec:FusionNet}).
$\mathbf{I}_i$ and $\mathbf{I}_c$ are the vehicle movement signals from the IMU and chassis. The $\mathbf{I}_i$ includes $3$-DOF acceleration and $3$-DOF angular velocity, while the $\mathbf{I}_c$ contains nominal vehicle speed, four-wheel speed, steering wheel angle, vehicle stationary signal, forward acceleration, lateral acceleration, and heading angular velocity.

The control variables $\mathbf{u}=[v, a, \omega]^T$ (i.e., vehicle velocity, vehicle acceleration, and turning rate) are refined to $\hat{\mathbf{u}}$ by applying $\delta \mathbf{u}$ as (\ref{eq:apply_res}):

\begin{equation}
\label{eq:apply_res}
\hat{\mathbf{u}} = \mathbf{u} + \delta \mathbf{u}.
\end{equation}

Unlike the learning-based odometry method \cite{esfahani2019aboldeepio} that predicts the magnitude of translation and rotation changes directly from a neural network, we opt to predict a small correction, denoted as $\delta \mathbf{u}$, to refine the control variables $\hat{\mathbf{u}}$. The $\hat{\mathbf{u}}$ is then used to predict the vehicle state by adhering to the rules of the conventional motion model, thereby maintaining stable vehicle movement. By predicting small corrections instead of the full magnitude, the learning task is simplified, and the risk of large prediction errors is reduced.
Then the state $\mathbf{X}^{\mathtt{pred}}_k$, representing the $k-$th state prediction, is calculated from the last predicted state $\mathbf{X}^{\mathtt{pred}}_{k-1}$ and the control variables from the $k$-th frame $\hat{\mathbf{u}}_k$, as shown in (\ref{eq:pred_ctra}):
\begin{equation}
\begin{gathered}
\mathbf{X}^{\mathtt{pred}}_{k} = \mathbf{X}^{\mathtt{pred}}_{k-1} + \mathbf{A}\cdot \hat{\mathbf{u}}_k \cdot dt \\
\mathbf{A} =
\begin{bmatrix}
\cos(\mathbf{X}^{\mathtt{pred}}_{k-1,\psi}) & \mathbf{0}_{1\times 2}  \\
\sin(\mathbf{X}^{\mathtt{pred}}_{k-1,\psi}) & \mathbf{0}_{1\times 2} \\
\mathbf{0}_{2\times 1} & \mathbf{I}_{2\times 2} \\
\end{bmatrix}
\label{eq:pred_ctra}
\end{gathered}
\end{equation}

\noindent where $dt$ denotes the time interval between the last predicted timestamp and the current timestamp.

The relative state change is supervised against the ground truth pose change, with the Huber loss function $\boldsymbol{\mathcal{H}}(\cdot)$ applied to reduce the influence of outliers. The relative state loss of $k$-th frame $d_k$ is indicated in (\ref{eq:loss_rel}):

\begin{equation}
\begin{gathered}
d_{k} = \boldsymbol{\alpha}  \cdot \boldsymbol{\mathcal{H}}(\boldsymbol{\pi}(\mathbf{X}^{\mathtt{pred}}_{k}, \mathbf{X}^{\mathtt{pred}}_{k-1})-
\boldsymbol{\pi}(\bar{\mathbf{X}}_{k}, \bar{\mathbf{X}}_{k-1})) \\
\boldsymbol{\pi}(\mathbf{X}_a, \mathbf{X}_b) = \left[ x_a - x_b,~ y_a - y_b,~ v_a - v_b,~ \phi(\psi_a, \psi_b) \right]^T
\label{eq:loss_rel}
\end{gathered}
\end{equation}

\noindent where $\bar{\mathbf{X}}_{k}$ and $\bar{\mathbf{X}}_{k-1}$ represents the ground truth state at the $k$-th and $(k-1)$-th frame. The vector $\boldsymbol{\alpha}$ is a $1 \times 4$ matrix that assigns weights to the different loss terms.
The function $\boldsymbol{\pi}(\cdot) $ is a $4\times 1$ vector that calculates the relative state change between two states $\mathbf{X}_a$ and $\mathbf{X}_b$. The operation $\phi(\cdot) \in [0, 2\pi)$ computes the angular difference.

\subsection{MeasurementNet}
The \textit{MeasurementNet} $\boldsymbol{\mathcal{N}}$ can be described as (\ref{eq:meas_net}):

\begin{equation}
\mathbf{R} = \boldsymbol{\mathcal{N}}([\mathbf{I}_{i},\mathbf{I}_{c},\mathbf{I}_{g}]) \label{eq:meas_net}
\end{equation}
\noindent where $\mathbf{R}$ denotes the output measurement feature.
$\mathbf{I}_{g}$ is the input feature comprising signals from GNSS receivers. As Guohao \textit{et al.} \cite{zhang2021prediction} demonstrated, predicting GNSS satellite visibility and pseudorange error based on GNSS measurement-level data can effectively reflect the quality of GNSS measurements. In our system, similar signals are used, including the number of tracked satellites, original measurement covariance, Dilution of Precision (DOP), and solution type.
Additionally, incorporating IMU $\mathbf{I}_{i}$ and chassis signals $\mathbf{I}_{c}$ helps validate the GNSS measurement quality by leveraging information from different modalities.

Since the measurement feature is closely linked to measurement quality, an auxiliary loss function $e$ is introduced to supervise the validity of GNSS position, velocity, and heading measurements. These measurement validities are defined by a position error less than \SI{1}{\meter}, a velocity error less than \SI{0.5}{\meter/\second}, and a heading error less than \SI{0.1}{\radian}. The auxiliary loss $e$ is given by (\ref{eq:loss_aux}):

\begin{equation}
\begin{gathered}
\mathbf{Y} = \boldsymbol{\mathcal{C}}(\mathbf{R}) \\
e = {\mathcal{E}}(\mathbf{Y},\bar{\mathbf{Y}}) \label{eq:loss_aux}
\end{gathered}
\end{equation}

\noindent where $\mathbf{Y} \in \mathbb{R}^{3 \times 1}$ is the estimated category scores of measurement validity computed by an additional neural network \textit{MeasurementClassifier} $\boldsymbol{\mathcal{C}}$, $\bar{\mathbf{Y}}$ is the ground truth label. ${\mathcal{E}}$ is the cross entropy loss.

\subsection{FusionNet}\label{sec:FusionNet}

\begin{figure}[t]
	\centering
	\includegraphics[width=1\linewidth]{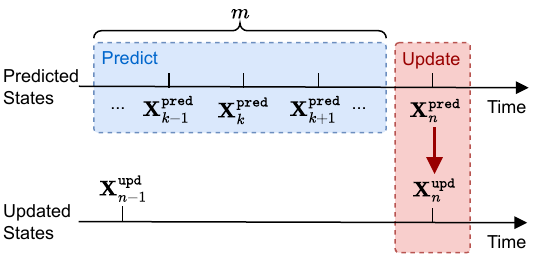}
	\caption{The relationship between predicted states and updated states.}
	\label{fig:states_dem}
	\vspace{-0.6cm}
\end{figure}

As illustrated in Fig.~\ref{fig:states_dem}, the predicted and updated states are asynchronous, with $m$ predicted states occurring between each pair of updated states. Given the differing indices, we use $k$ and $n$ to denote the index of the predicted and updated state respectively. Specifically, $\mathbf{X}^{\mathtt{pred}}_{k}$ represents the $k$-th predicted state, while $\mathbf{X}^{\mathtt{upd}}_{n}$ denotes the $n$-th updated state.

\textit{FusionNet} $\boldsymbol{\mathcal{F}}$, an RNN, is described in (\ref{eq:fusion_net}):

\begin{equation}
\begin{gathered}
\mathbf{O_n} = \boldsymbol{\mathcal{F}}(\mathbf{R}_{n}, \mathcal{Q}, \tilde{\mathbf{r}}_{n}, \mathbf{V}_{n-1}) \\
\mathcal{Q} =[\mathbf{Q}^{k-m+1}, \mathbf{Q}^{k-m+2}, ..., \mathbf{Q}^{k}]
\label{eq:fusion_net}
\end{gathered}
\end{equation}
\noindent where $\mathbf{O_n}=[\mathbf{o}_n, \mathbf{V}_{n}]$ consists the network output $\mathbf{o}_{n} \in \mathbb{R}^{4 \times 1}$ and the hidden state $\mathbf{V}_{n}$. $\mathbf{R}_{n}$ denotes the measurement feature from the $n$-th frame, and $\mathcal{Q}$ represents the concatenated historical motion features with a window size $m$.
Additionally, to incorporate an outlier measurement rejection mechanism, the measurement innovations $\tilde{\mathbf{r}}_{n}$ are also fed to $\boldsymbol{\mathcal{F}}$, which is calculated as (\ref{eq:inno}):

\begin{equation}
\begin{gathered}
\mathbf{r}_{n} = \mathbf{z}_{n} - \mathbf{H}_{n} \mathbf{X}^{\mathtt{pred}}_{n} \\
\tilde{\mathbf{r}}_{n} = \mathbf{tanh}\left(\frac{\mathbf{r}_n}{\mathbf{c}}\right)
\label{eq:inno}
\end{gathered}
\end{equation}

\noindent where $\mathbf{r}_{n}$ denotes the original measurement innovation, $\mathbf{z}_{n}$ represents the measurement, and $\mathbf{H}_{n}$ is the observation matrix.
$\tilde{\mathbf{r}}_{n}$ is the normalized measurement innovation, and $\mathbf{c}$ is a constant scalar vector used for scaling the states. Measurement residuals are normalized using the $\mathbf{tanh}(\cdot)$ function to manage extreme values effectively.

Similar to \textit{KalmanNet} \cite{revach2022kalmannet}, which estimates the Kalman filter gain without the noise covariance matrix,
the network output $\mathbf{o}_{n}$ is used for measurement fusion by applying sigmoid function $\sigma (\cdot)$ to obtain the estimated residual weight $\mathbf{w}_{n}$, which is bounded within the range $[0, 1]$. Given that we have a full-state measurement residual, element-wise multiplication is performed on the residual vector to update the state, as shown in (\ref{eq:fuse}):

\begin{equation}
\begin{gathered}
\mathbf{w}_n = \sigma(\mathbf{o}_n) \\
\mathbf{X}^{\mathtt{upd}}_{n} = \mathbf{X}^{\mathtt{pred}}_{n} + \mathbf{w}_n \odot \mathbf{r}_n
\label{eq:fuse}
\end{gathered}
\end{equation}
\noindent where $\odot$ denotes element-wise multiplication.

The updated state $\mathbf{X}^{\mathtt{upd}}_{n}$ is supervised with its corresponding ground truth $\bar{\mathbf{X}}_{n}$ as demonstrated in (\ref{eq:loss_abs}):

\begin{equation}
f_n = \boldsymbol{\beta} \cdot \boldsymbol{\mathcal{H}}(\boldsymbol{\pi}(\mathbf{X}^{\mathtt{upd}}_{n}, \bar{\mathbf{X}}_{n})) \label{eq:loss_abs}
\end{equation}
\noindent where $\boldsymbol{\beta}$ is a $1 \times 4$ matrix of loss term weights
ensuring better loss balance. The term $f_n$ denotes the absolute pose loss for the $n$-th frame.

The total loss $l$ is the sum of the losses $d$, $e$, and $f$ from all the frames, as shown in  (\ref{eq:total_loss}):

\begin{equation}
l = \sum^{K}_{k=0}d_k + \sum^{N}_{n=0}e_n + \sum^{N}_{n=0}f_n
\label{eq:total_loss}
\end{equation}
where $K$ is the number of predicted states, $N$ is the number of updated states.

\section{Experiment}

\begin{figure}[t]
	\centering
	\includegraphics[width=0.99\linewidth]{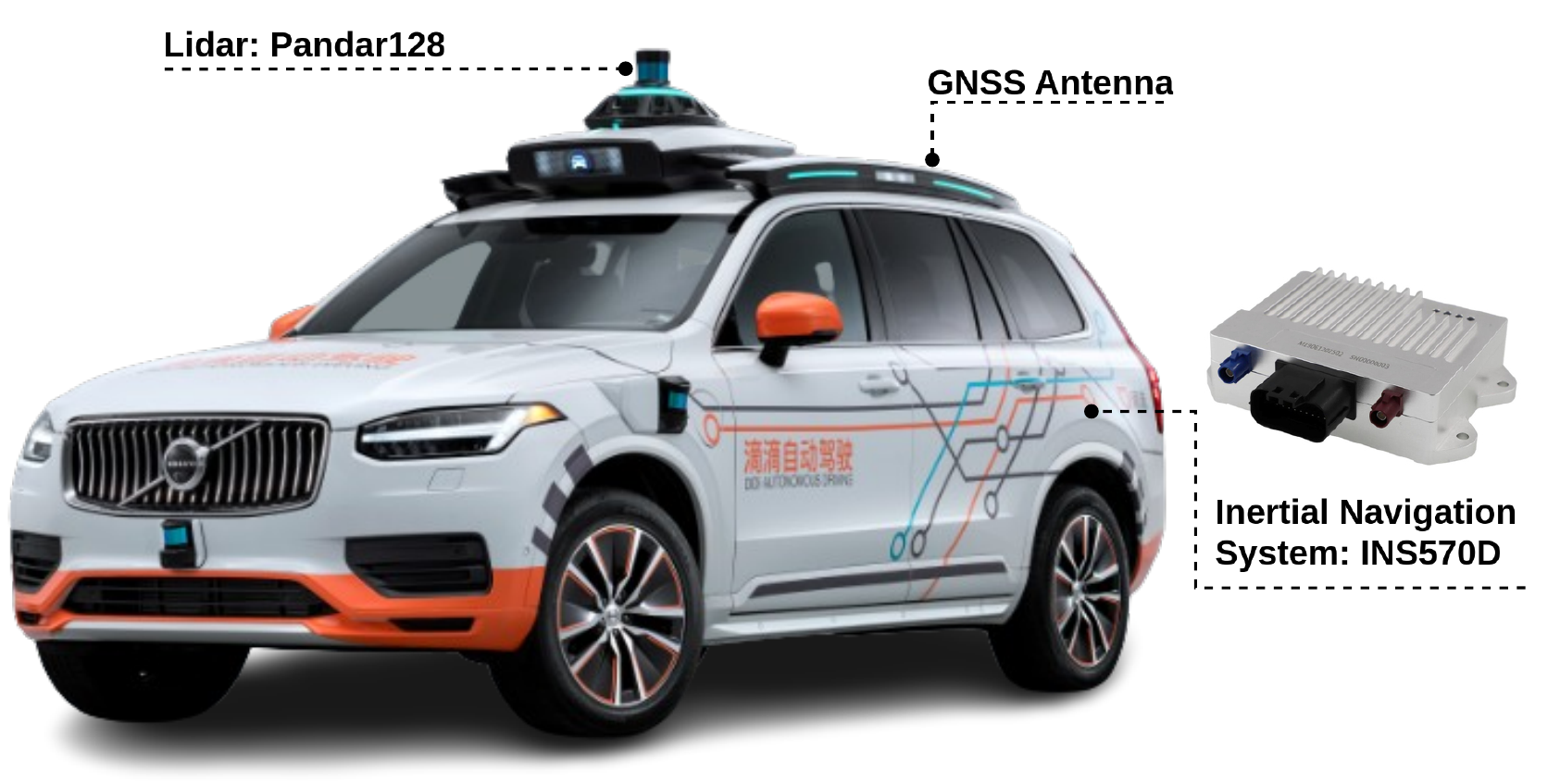}
	\caption{Our experimental vehicle is equipped with a $128$-line LiDAR, a dual-antenna GNSS receiver, and an IMU.}
	\label{fig:car}
	\vspace{-0.5cm}
\end{figure}

\subsection{Experimental Setup}
\begin{figure*}[h]
\centering
	\subfloat[Urban canyon]{
		\includegraphics[width=0.3\textwidth]{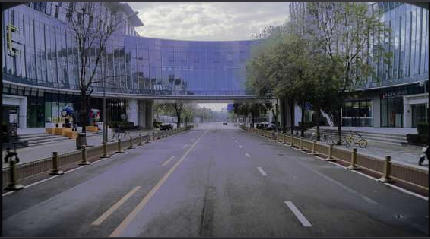}
		\label{fig:subfig1}
	}
	\subfloat[Tunnel]{
		\includegraphics[width=0.3\textwidth]{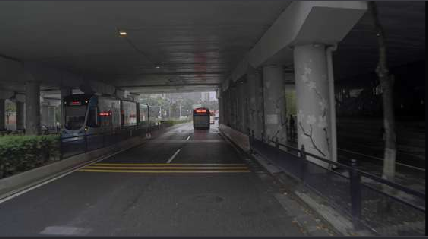}
		\label{fig:subfig2}
	}
	\subfloat[Dense Canopy]{
		\includegraphics[width=0.3\textwidth]{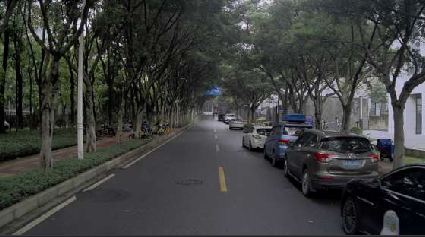}
		\label{fig:subfig3}
	}

	\caption{Some challenging environments for GNSS localization in our autonomous driving dataset.}
	\label{fig:real_scene}
        \vspace{-0.5cm}
\end{figure*}

We conducted experiments using a real-world autonomous driving dataset collected in urban areas of China with the autonomous driving vehicle depicted in Fig.~\ref{fig:car}.
The dataset encompasses a range of driving scenarios, including U-turns, left-turns, right-turns, roundabouts, and more. It also features challenging environments for GNSS localization, such as dense canopy areas, urban canyons, and tunnels, as shown in Fig.~\ref{fig:real_scene}.
The IMU and chassis signals are sampled at \SI{100}{\hertz}, while GNSS measurements are recorded at \SI{5}{\hertz}.
The ground truth pose is generated using a LiDAR-based SLAM algorithm that leverages LOAM \cite{zhang2014loam} for scan matching and odometry estimation, while pose graph optimization \cite{dellaert2017factor} is applied for global consistency.

For training, the data is segmented into 20-second and 5-second segments. The shorter 5-second segments are portions with poor GNSS data to achieve a better balance between poor and good samples.
For testing, a $20$-second-segments test dataset is used to cover various scenarios and vehicle movements, while a \textit{trail trip} test dataset lasting \SI{29.2}{\min} and covering \SI{10.6}{km} is employed to evaluate long-term performance.
Details of the datasets are provided in Table \ref{tab:dataset_table}.

\begin{table}[h]
\caption{Dataset Attributes}
\centering
\begin{tabular}{lcc}
\toprule
\textbf{Dataset} & \textbf{Segment Duration (\SI{}{s})} & \textbf{Segment Number} \\
\midrule
Train-short      & $5$ & $10,719$ \\
Train-long       & $20$ & $18,869$ \\
Test       & $20$ & $1,062$ \\
\bottomrule
\end{tabular}
\label{tab:dataset_table}
\vspace{-0.4cm}
\end{table}

For the neural network structure, \textit{MotionNet} is an MLP with seven hidden layers, where the motion feature is extracted from the fifth layer, producing a 64-dimensional feature map. \textit{MeasurementNet} is an RNN that includes an MLP with four linear layers, followed by an LSTM with a 64-dimensional hidden state, resulting in a 64-dimensional measurement feature. \textit{FusionNet} is an RNN consisting of a single-layer MLP encoder, followed by an LSTM and a three-layer MLP decoder. The \textit{MeasurementClassifier} is an MLP with two linear layers. All dropout layers have a rate of $0.1$.

The training process is conducted in two stages.
In the first stage, all the training data is divided into \SI{5}{\second} segments to ensure stable convergence and prevent gradient explosion, and the model is trained for $3000$ epochs.
In the second stage, we fine-tune the model for $600$ epochs using \SI{20}{\second} segments, enabling the network to refine its fusion strategy with extended historical data.
The model is optimized using the AdamW optimizer with a $0.01$ initial learning rate and a batch size of $8192$. The loss weights $\boldsymbol{\alpha}$ and $\boldsymbol{\beta}$ are set to $[1\text{e}3, 1\text{e}3, 1\text{e}3, 1\text{e}5]$ and $[1, 1, 1, 10]$, respectively.
The entire system is implemented using \textit{PyTorch} and executed on a \textit{NVIDIA A6000 GPU} and an \textit{Intel Xeon Platinum 8352Y CPU}. 

\begin{figure}[!t]
	\centering
	\subfloat[The distribution of position error.]{
		\includegraphics[width=0.48\textwidth]{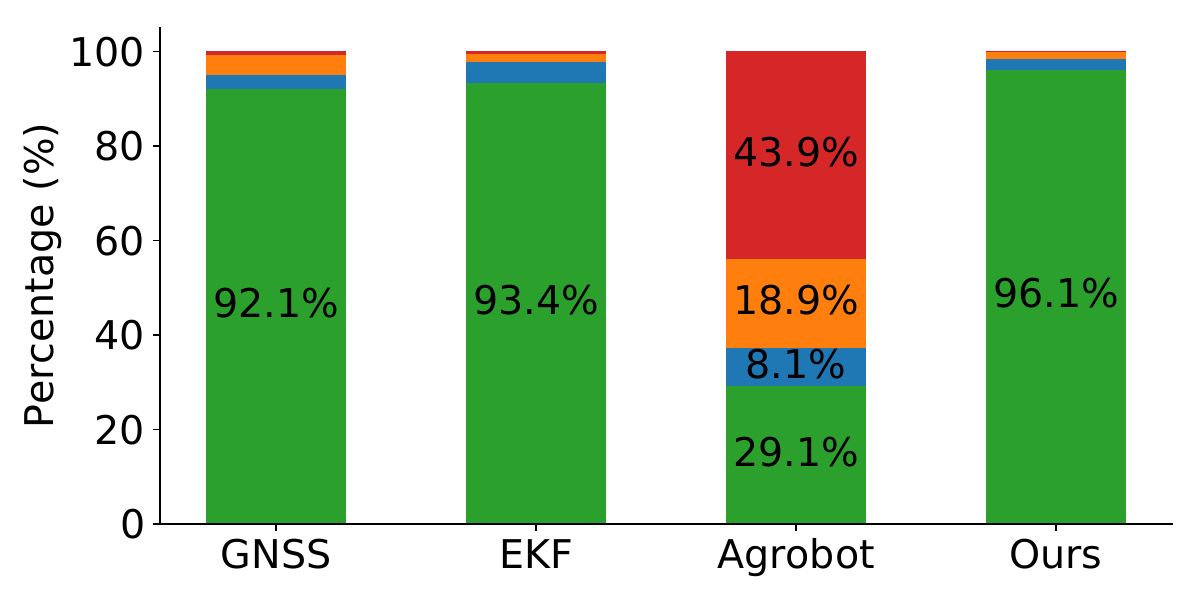}
		\label{fig:hist1}
	}\\
        \vspace{-0.3cm}
	\subfloat[The distribution of heading error.]{
		\includegraphics[width=0.45\textwidth]{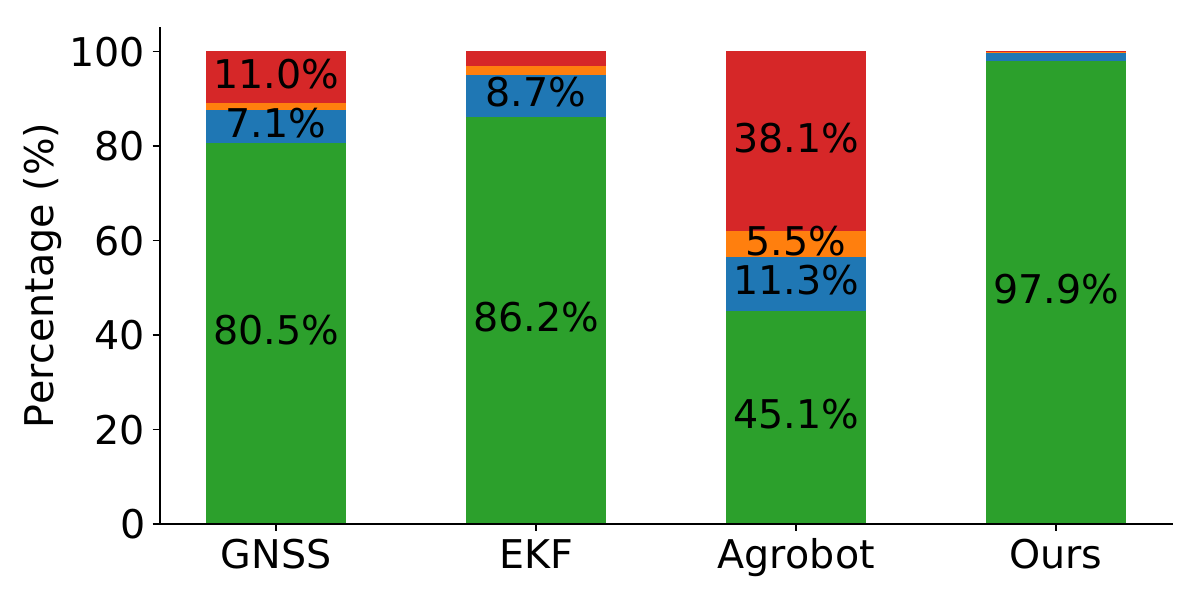}
		\label{fig:hist2}
	}	
	\caption{Stacked column charts of error distributions: (a) Position Error and (b) Heading Error.}
	\label{fig:combined_histograms}
	\vspace{-0.50cm}
\end{figure}

\subsection{Quantitative Evaluation}

\begin{table*}[h]
\caption{RMSE metrics for different methods.}
\centering
\setlength{\tabcolsep}{4pt} 
\begin{tabular}{lcccccc}
\toprule
\multirow{2}{*}{} & \multicolumn{3}{c}{$20$-second-segments} & \multicolumn{3}{c}{\textit{Trail Trip}} \\
\cmidrule(lr){2-4} \cmidrule(lr){5-7}
 & Position (\si{\meter}) & Velocity (\si{m/s}) & Heading (\si{\radian}) & Position (\si{\meter}) & Velocity (\si{m/s}) & Heading (\si{\radian}) \\
\midrule
GNSS & $1592.9497$ & $0.3140$ & $0.6208$ & $2770.3798$ & $0.1319$ & $0.3277$ \\
EKF & $0.4971$    & $0.1044$ & $0.0182$ & $0.4877$  & $0.1323$ & $\mathbf{0.0098}$ \\
\textit{Agrobot} & $5.5831$ &  $0.3557$ & --- & $9.1146$ & $0.2237$ & --- \\
Ours($@$5.0Hz, Proposed) & $\mathbf{0.2718}$ & $\mathbf{0.1031}$ & $0.0094$ & $0.4154$  & $\mathbf{0.0866}$ & $0.0124$ \\
Ours($@$2.5Hz) & $0.2966$ & $0.1494$ & $\mathbf{0.0079}$ & $\mathbf{0.3554}$ & $0.1247$ & $0.0178$ \\
Ours($@$1.0Hz) & $0.5348$ & $0.2545$ & $0.0099$ & $0.5800$ & $0.2352$ & $0.0283$ \\
\bottomrule
\end{tabular}
\label{tab:rmse_comparison}
\vspace{-0.5cm}
\end{table*}

\begin{figure*}[h]
	\centering
	\subfloat[Poor GNSS]{
		\includegraphics[width=0.3\textwidth]{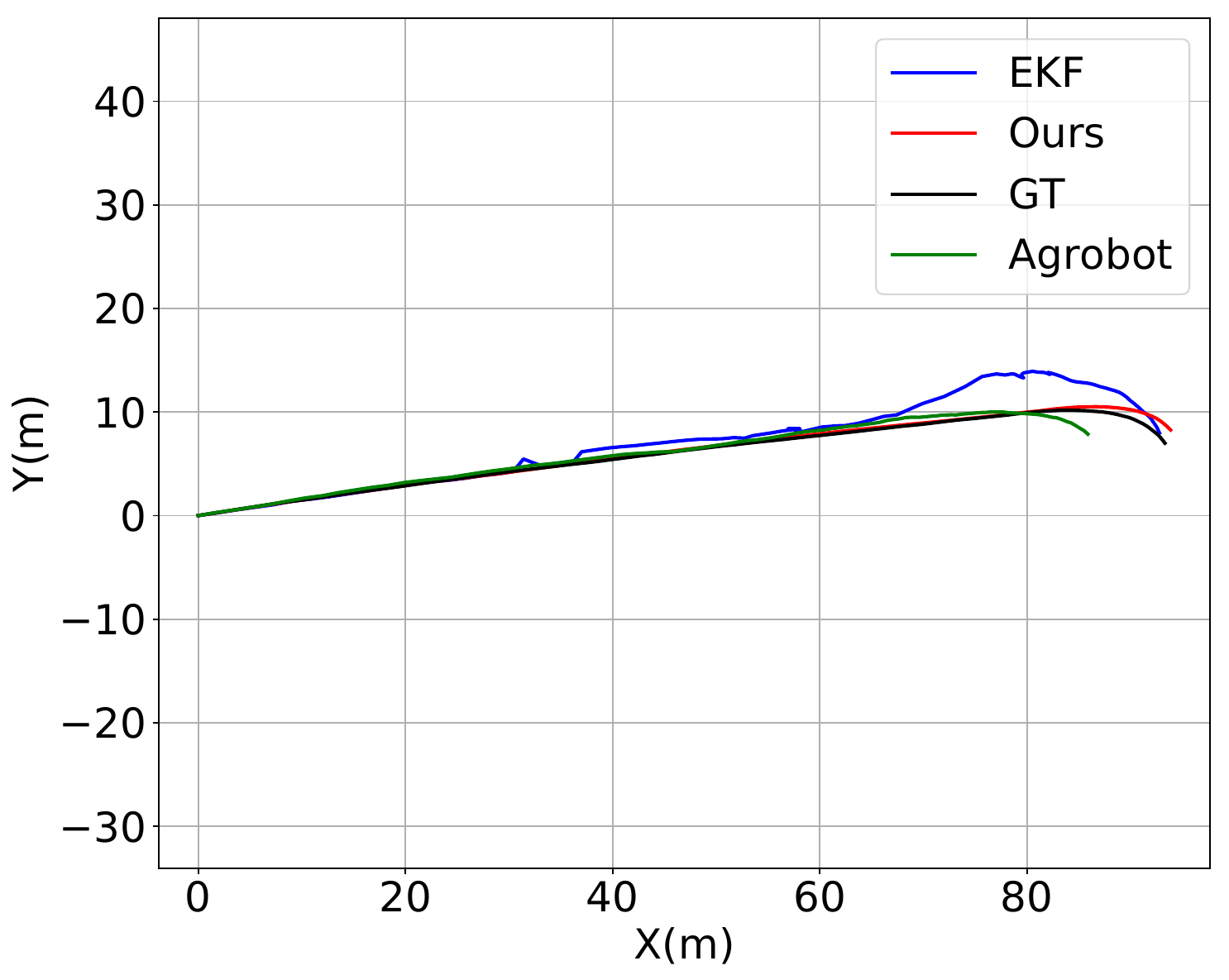}
		\label{fig:subfig1}
	}
	\subfloat[Right turn]{
		\includegraphics[width=0.3\textwidth]{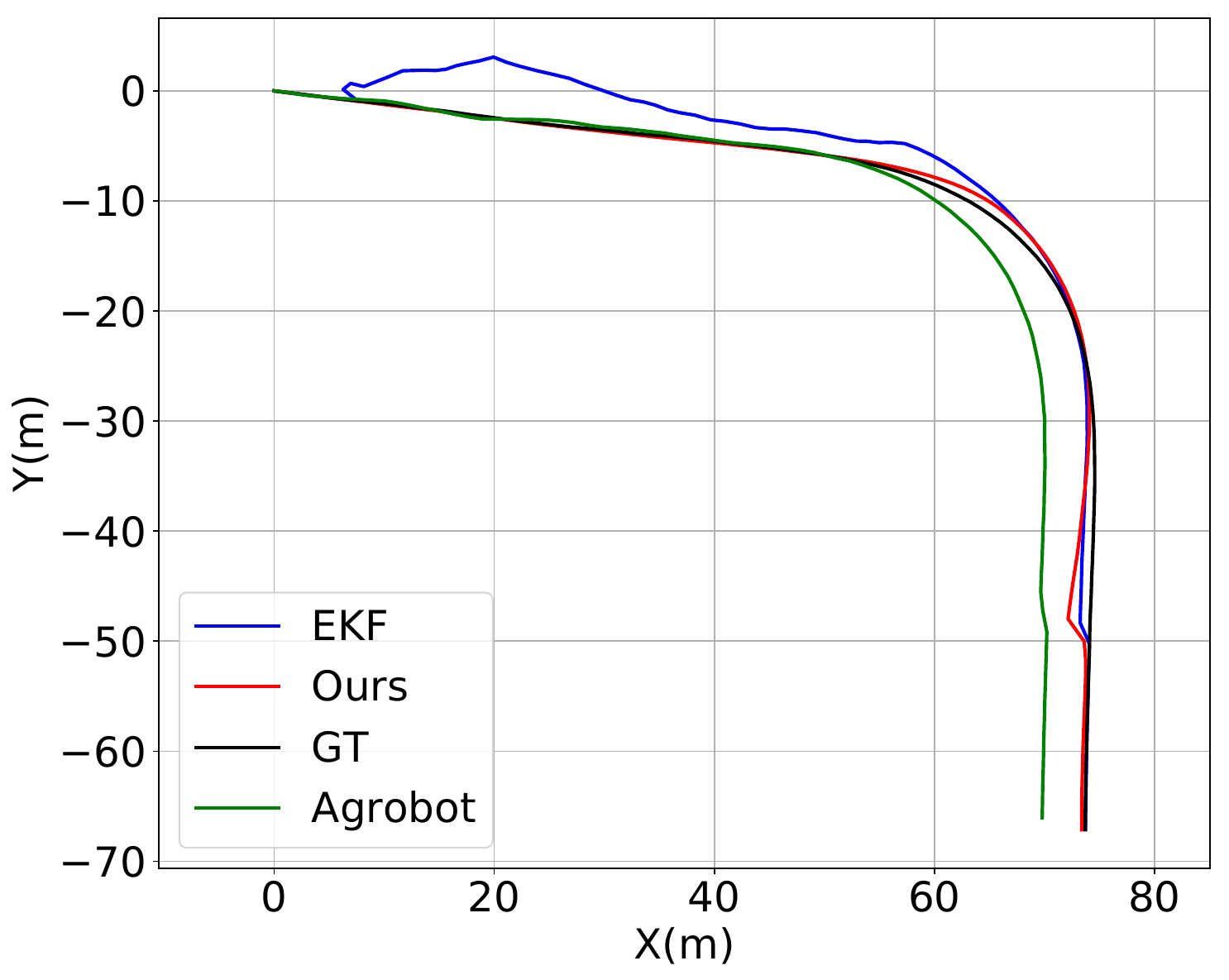}
		\label{fig:subfig2}
	}
	\subfloat[Moving forward with nudging]{
		\includegraphics[width=0.3\textwidth]{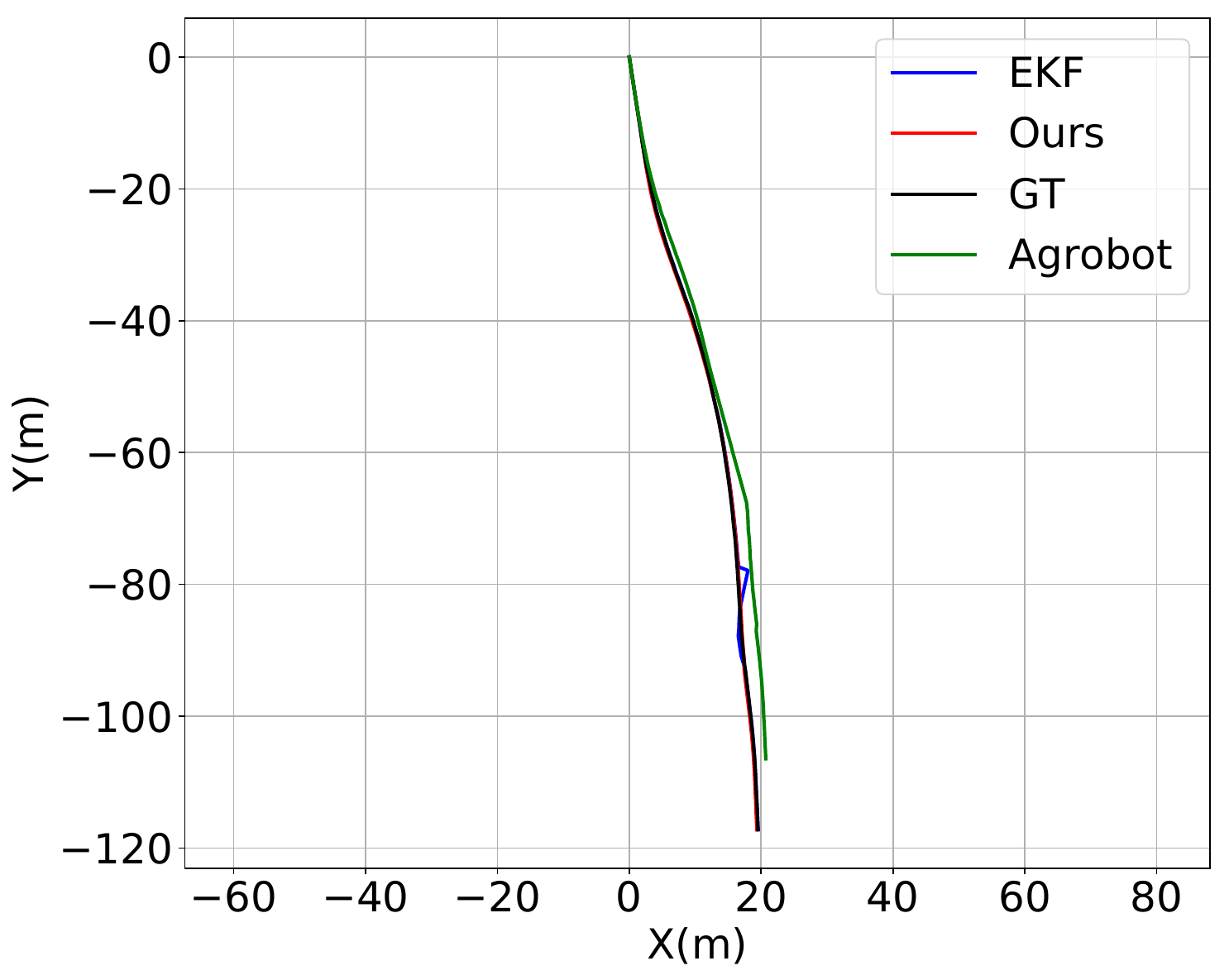}
		\label{fig:subfig3}
	}
	\\
        \vspace{0.05cm}
	\subfloat[Moving forward in high speed]{
		\includegraphics[width=0.3\textwidth]{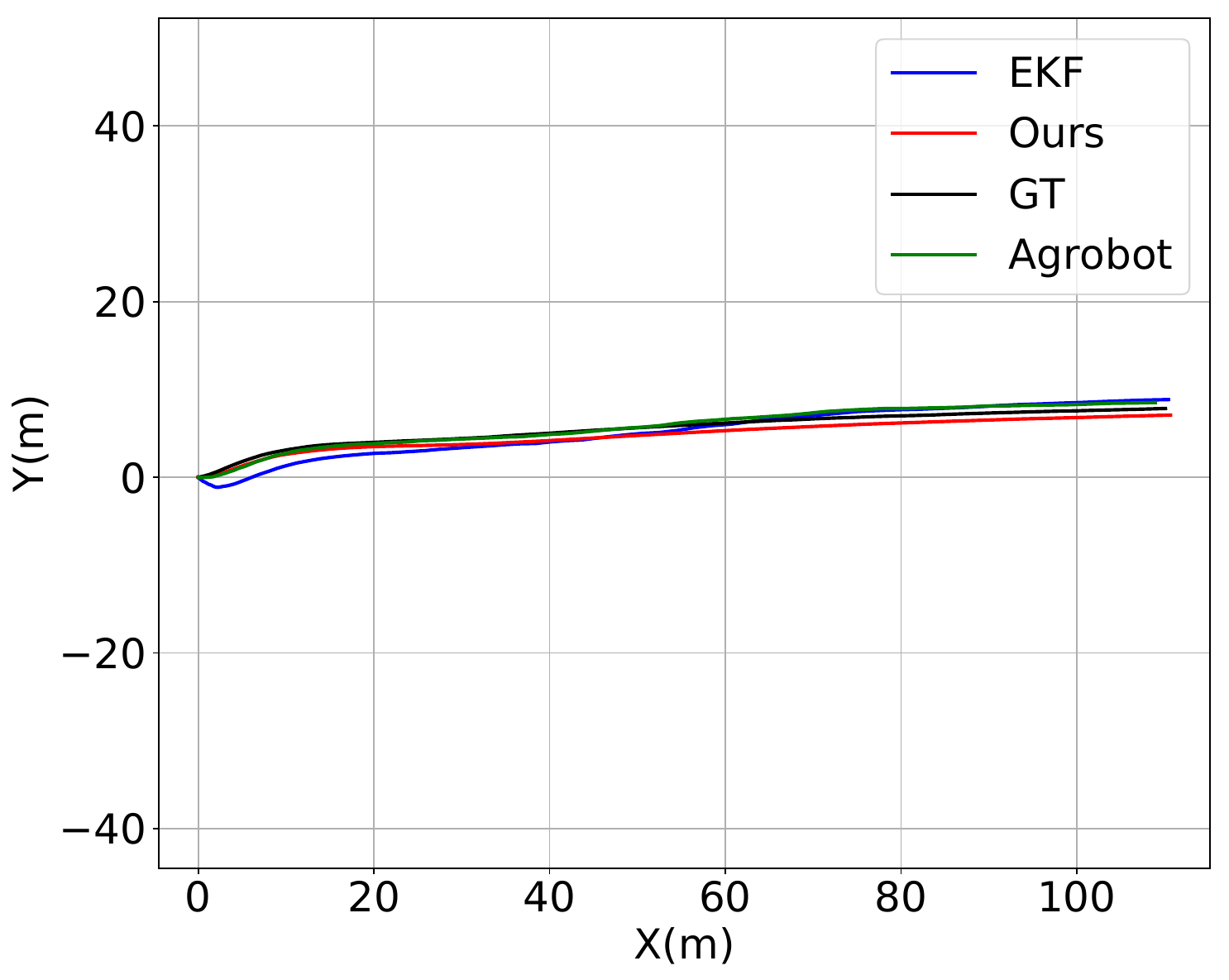}
		\label{fig:subfig4}
	}
	\subfloat[Slowing moving with poor GNSS quality]{
		\includegraphics[width=0.3\textwidth]{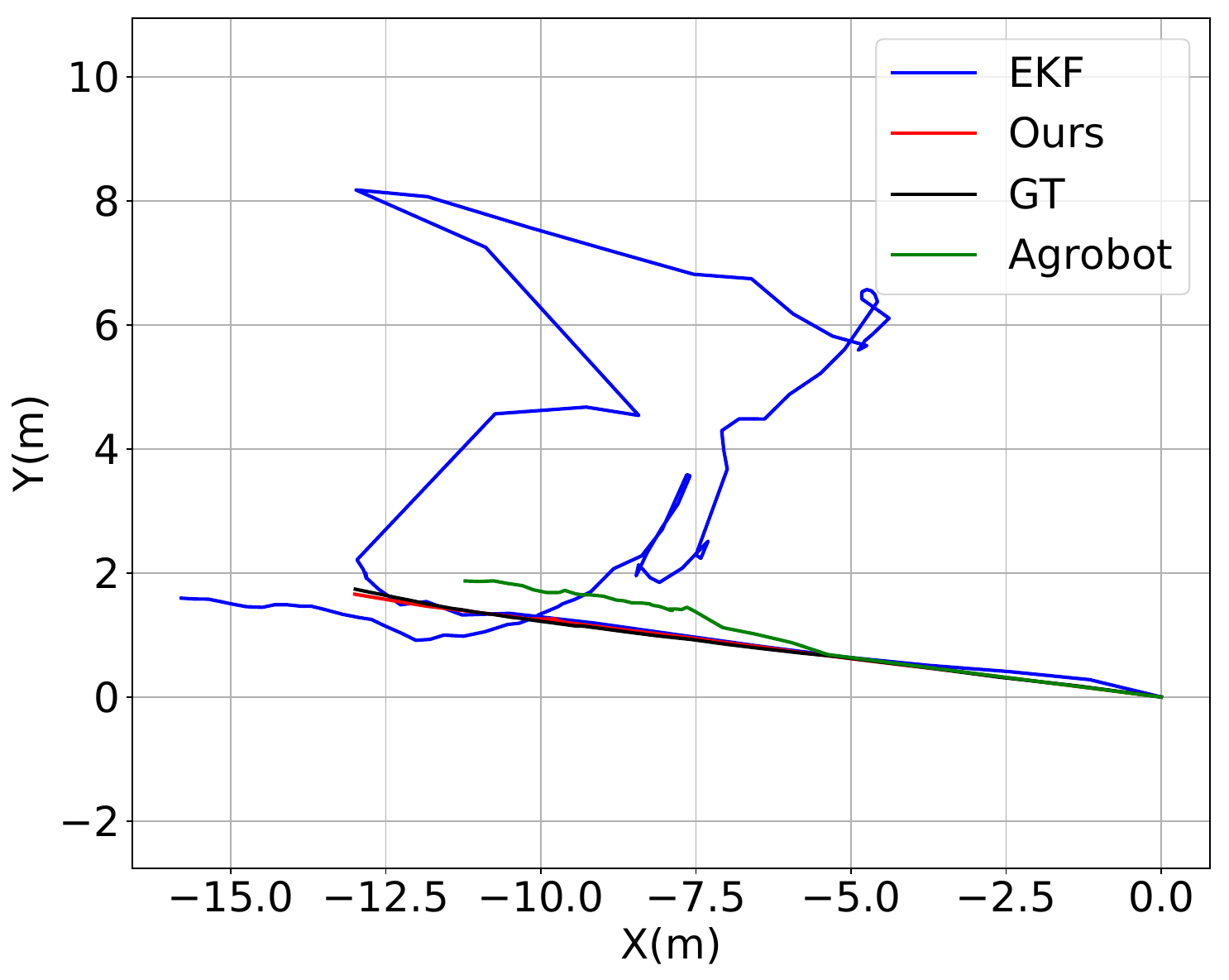}
		\label{fig:subfig5}
	}
	\subfloat[Slowing moving with poor GNSS quality]{
		\includegraphics[width=0.3\textwidth]{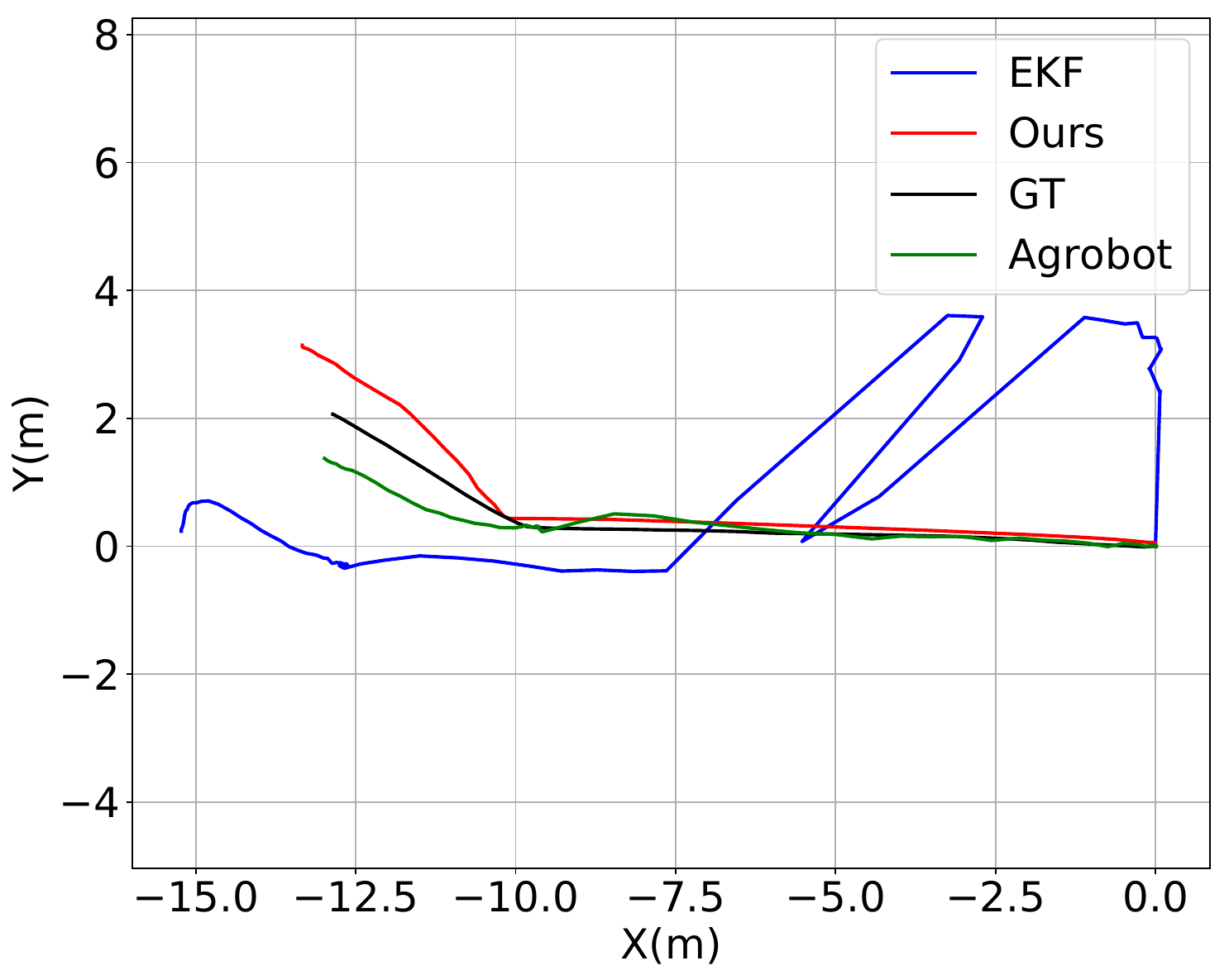}
		\label{fig:subfig6}
	}
	\caption{Our method achieves more stable pose estimation across various test scenarios compared to both the EKF and \textit{Agrobot}.}
	\label{fig:short_seg}
        \vspace{-0.6cm}
	\centering
	
\end{figure*}

We compare our method with three other approaches: the original GNSS measurment, the traditional EKF and the neural network-based GNSS/IMU algorithm, \textit{Agrobot} \cite{Du2023Agrobot}.
The GNSS measurment provided by a GNSS receiver with dual-antenna can provide position, velocity and heading measurement.
The EKF utilizes GNSS measurements for updates and predictions using IMU and chassis signals. The process noise covariance matrix of the EKF is tuned with fixed values, while the measurement noise covariance is adjusted according to the GNSS measurement covariance provided by the GNSS receiver.
\textit{Agrobot} uses megnetometer to provide a global heading measurment, but our vehicle lacks a magnetometer, we adapted the comparison by converting the GNSS heading into a magnetometer vector, allowing us to train the model on our dataset. In this method, a sliding window of size $100$ frames is used to process the IMU sequence, with a stride of $20$ frames, corresponding to a duration of \SI{200}{\milli\second}. This implies that, unlike our method, which performs one update with $20$ predictions, \textit{Agrobot} performs one update with one prediction.
All methods, except for GNSS measurements, are initialized with the ground truth pose at the start of the test sequence.

We evaluate the performance of these methods using Root Mean Square Error (RMSE) for position, velocity, and heading on both the $20$-second-segments test dataset and the \textit{trail trip} test dataset. Since the \textit{Agrobot} only estimates position and velocity, its heading is not evaluated, as summarized in Table \ref{tab:rmse_comparison}.
The error distribution for position and heading across all methods is computed for the $20$-second-segments test dataset, with the results presented in the stacked column charts in Fig. \ref{fig:combined_histograms}.

\begin{figure*}
    \centering
    \includegraphics[width=\textwidth]{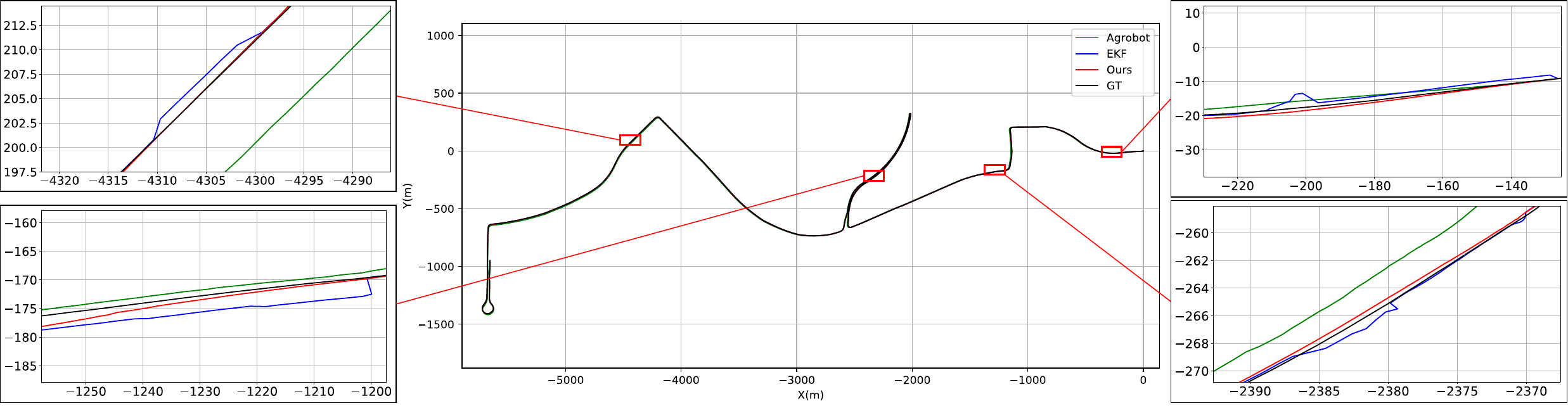}
    \caption{Trajectory of the \textit{trail trip} with zoomed-in sections highlighting key scenarios, demonstrating the superior stability and accuracy of our method (in red) under poor observation conditions.}
    \label{fig:long_seg}
    \vspace{-0.6cm}
\end{figure*}

As demonstrated by the RMSE metrics and error distribution, our method shows superior performance in estimating position, velocity, and heading. The GNSS measurements suffer from significant outliers caused by adverse conditions, resulting in large errors. The EKF, which employs a fusion approach, achieves better results than raw GNSS measurements by reducing the distribution of large errors.
However, the EKF's reliance on the measurement covariance matrix provided by the GNSS receiver can be problematic, as this covariance is often inaccurately estimated, particularly in conditions of poor GNSS visibility.
Our method surpasses the EKF by dynamically adjusting both the measurement and motion features, allowing for a more accurate assessment of GNSS measurement reliability and yielding a more stable pose output. While most metrics show improvements over other methods, a slight discrepancy in heading accuracy on the \textit{trail trip} compared to the EKF was observed. This is due to limited variation in the heading during the $20$-second-segments training dataset, resulting in insufficient training for heading estimation.
In contrast, the \textit{Agrobot} performs worse than the EKF because its velocity estimation, based on IMU data without a base velocity reference, leads to considerable position and velocity errors.
Although the \textit{Agrobot} is optimized for low-speed, short-range movements typical in agricultural fields and relies on fixed patterns learned during training, this approach proves less effective for road vehicles that encounter dynamic maneuvers.
Consequently, the positioning performance reported in their study \cite{Du2023Agrobot} is notably better on seen trajectories (training dataset) compared to unseen trajectories (test dataset).

In addition, we conducted ablation experiments by varying the update frequency to \SI{2.5}{Hz} and \SI{1.0}{Hz}, despite the model being trained at \SI{5.0}{Hz}. The results demonstrate the algorithm's generalization and robustness, as it maintains strong performance even at lower update frequencies. Notably, at \SI{2.5}{Hz}, the heading RMSE on the 20-second segment test dataset and the position RMSE on the \textit{trail trip} test dataset outperform those at \SI{5.0}{Hz}. This improvement can be attributed to two key factors: first, reducing GNSS measurements helps mitigate measurement noise in GNSS data in some cases; second, the algorithm effectively compensates for fewer GNSS updates by leveraging \textit{MotionNet} predictions. Overall, as shown in Table \ref{tab:rmse_comparison}, our proposed method operating at an update frequency of \SI{5.0}{Hz} achieves the best performance across most metrics.

We also assessed the inference time across different methods. Noting the varying ratios of prediction to update stages, specifically, our method and the EKF perform an update after every $20$ predictions, while the \textit{Agrobot} method performs an update after each prediction. We calculated the total algorithm runtime between consecutive updates to ensure a fair comparison.
Both our method and the \textit{Agrobot} were executed on a GPU, with execution times of \SI{8.70}{ms} and \SI{89.77}{ms}, respectively. The EKF was implemented on both GPU and CPU platforms. The CPU implementation of the EKF achieved the fastest execution time at \SI{4.30}{ms}, while the \textit{EKF-GPU} implementation demonstrated slower performance at \SI{14.19}{ms} due to differences in computational architecture. The \textit{Agrobot} exhibited the longest runtime due to its use of sliding windows for processing historical data at each inference.
In contrast, our method utilizes hidden states to retain historical information, processing only the current frame. This makes it particularly well-suited for real-time applications such as autonomous driving, with minimal regression compared to the EKF. 

\subsection{Qualitative Evaluation}
We present representative scenarios from our test dataset. Fig.~\ref{fig:short_seg} illustrates six typical cases from the $20$-second-segments test dataset, including various movements such as straight motion, nudging, slow movement, and challenging scenarios involving GNSS signal loss due to signal blockage. Our method demonstrates more stable pose estimation compared to both the EKF and the \textit{Agrobot}.

Additionally, we evaluate our method over the comprehensive trip lasting \SI{29.2}{\min} and covering a distance of \SI{10.6}{\kilo\meter}. This trip encompasses a range of urban scenarios, including tunnels, left turns, right turns, U-turns, and roundabouts. Four challenging subsequences are highlighted in Fig.~\ref{fig:long_seg}, with their zoomed-in details presented in the corresponding subfigures. A video of the experiment can be viewed at \url{\videolink}. Our method demonstrates superior performance. It effectively mitigates the impact of poor GNSS measurements and adjusts the vehicle state appropriately when GNSS measurements are recovered and the vehicle's predicted state is suboptimal.

\section{Conclusion}

In this paper, we proposed a novel end-to-end neural network-based architecture for multi-sensor fusion in autonomous vehicle localization. Our approach encodes sensor data into feature vectors using neural networks, eliminating the need for parameter fine-tuning and proving the effectiveness of neural network-based fusion. Our extensive experimentation on a challenging urban autonomous driving dataset confirms that our method outperforms the existing methods in terms of accuracy and robustness, highlighting the potential of learning-based multi-sensor fusion approaches.

However, our approach has certain limitations. Notably, the vehicle state is simplified to a 2D planar model, which does not account for road slope variations. Future work will focus on extending this model to a $6$-DOF localization system. Additionally, we plan to integrate additional sensor modalities, such as LiDAR and cameras, into the multi-sensor fusion framework to further enhance localization accuracy and robustness across diverse driving conditions.

\bibliographystyle{IEEEtran}
\bibliography{ref}

\begin{thebibliography}{10}
\providecommand{\url}[1]{#1}
\csname url@rmstyle\endcsname
\providecommand{\newblock}{\relax}
\providecommand{\bibinfo}[2]{#2}
\providecommand\BIBentrySTDinterwordspacing{\spaceskip=0pt\relax}
\providecommand\BIBentryALTinterwordstretchfactor{4}
\providecommand\BIBentryALTinterwordspacing{\spaceskip=\fontdimen2\font plus
\BIBentryALTinterwordstretchfactor\fontdimen3\font minus \fontdimen4\font\relax}
\providecommand\BIBforeignlanguage[2]{{%
\expandafter\ifx\csname l@#1\endcsname\relax
\typeout{** WARNING: IEEEtran.bst: No hyphenation pattern has been}%
\typeout{** loaded for the language `#1'. Using the pattern for}%
\typeout{** the default language instead.}%
\else
\language=\csname l@#1\endcsname
\fi
#2}}

\bibitem{hofmann2007gnss}
B.~Hofmann-Wellenhof, H.~Lichtenegger, and E.~Wasle, \emph{GNSS--global navigation satellite systems: GPS, GLONASS, Galileo, and more}.\hskip 1em plus 0.5em minus 0.4em\relax Springer Science \& Business Media, 2007.

\bibitem{Kalman_filter}
G.~Welch, G.~Bishop, \emph{et~al.}, ``An introduction to the kalman filter,'' 1995.

\bibitem{Particle_filter}
D.~Fox, S.~Thrun, W.~Burgard, and F.~Dellaert, ``Particle filters for mobile robot localization,'' in \emph{Sequential Monte Carlo methods in practice}.\hskip 1em plus 0.5em minus 0.4em\relax Springer, 2001, pp. 401--428.

\bibitem{dellaert2017factor}
F.~Dellaert, M.~Kaess, \emph{et~al.}, ``Factor graphs for robot perception,'' \emph{Foundations and Trends{\textregistered} in Robotics}, vol.~6, no. 1-2, pp. 1--139, 2017.

\bibitem{EKF}
\BIBentryALTinterwordspacing
S.~J. Julier and J.~K. Uhlmann, ``New extension of the kalman filter to nonlinear systems,'' in \emph{Defense, Security, and Sensing}, 1997. [Online]. Available: \url{https://api.semanticscholar.org/CorpusID:7937456}
\BIBentrySTDinterwordspacing

\bibitem{innovation_based_Adaptive_KF}
\BIBentryALTinterwordspacing
S.~Akhlaghi, N.~Zhou, and Z.~Huang, ``Adaptive adjustment of noise covariance in kalman filter for dynamic state estimation,'' \emph{2017 IEEE Power \& Energy Society General Meeting}, pp. 1--5, 2017. [Online]. Available: \url{https://api.semanticscholar.org/CorpusID:8409383}
\BIBentrySTDinterwordspacing

\bibitem{feng2014kalman}
B.~Feng, M.~Fu, H.~Ma, Y.~Xia, and B.~Wang, ``Kalman filter with recursive covariance estimation—sequentially estimating process noise covariance,'' \emph{IEEE Transactions on Industrial Electronics}, vol.~61, no.~11, pp. 6253--6263, 2014.

\bibitem{CELLO-3D}
\BIBentryALTinterwordspacing
D.~Landry, F.~Pomerleau, and P.~Gigu{\`e}re, ``Cello-3d: Estimating the covariance of icp in the real world,'' \emph{2019 International Conference on Robotics and Automation (ICRA)}, pp. 8190--8196, 2018. [Online]. Available: \url{https://api.semanticscholar.org/CorpusID:52916396}
\BIBentrySTDinterwordspacing

\bibitem{DICE}
\BIBentryALTinterwordspacing
K.~Liu, K.~Ok, W.~Vega-Brown, and N.~Roy, ``Deep inference for covariance estimation: Learning gaussian noise models for state estimation,'' \emph{2018 IEEE International Conference on Robotics and Automation (ICRA)}, pp. 1436--1443, 2018. [Online]. Available: \url{https://api.semanticscholar.org/CorpusID:52290152}
\BIBentrySTDinterwordspacing

\bibitem{Backprop-kf}
\BIBentryALTinterwordspacing
T.~Haarnoja, A.~Ajay, S.~Levine, and P.~Abbeel, ``Backprop kf: Learning discriminative deterministic state estimators,'' in \emph{Neural Information Processing Systems}, 2016. [Online]. Available: \url{https://api.semanticscholar.org/CorpusID:5670914}
\BIBentrySTDinterwordspacing

\bibitem{Multivariate_Uncertainty}
\BIBentryALTinterwordspacing
R.~L. Russell and C.~P. Reale, ``Multivariate uncertainty in deep learning,'' \emph{IEEE Transactions on Neural Networks and Learning Systems}, vol.~33, pp. 7937--7943, 2019. [Online]. Available: \url{https://api.semanticscholar.org/CorpusID:207757441}
\BIBentrySTDinterwordspacing

\bibitem{NLOS}
\BIBentryALTinterwordspacing
H.~Zhang, Z.~Wang, and H.~Vallery, ``Learning-based nlos detection and uncertainty prediction of gnss observations with transformer-enhanced lstm network,'' \emph{2023 IEEE 26th International Conference on Intelligent Transportation Systems (ITSC)}, pp. 910--917, 2023. [Online]. Available: \url{https://api.semanticscholar.org/CorpusID:261494052}
\BIBentrySTDinterwordspacing

\bibitem{Deep_RTK}
\BIBentryALTinterwordspacing
H.~Kim and T.~Bae, ``Deep learning-based gnss network-based real-time kinematic improvement for autonomous ground vehicle navigation,'' \emph{J. Sensors}, vol. 2019, pp. 3\,737\,265:1--3\,737\,265:8, 2019. [Online]. Available: \url{https://api.semanticscholar.org/CorpusID:109938265}
\BIBentrySTDinterwordspacing

\bibitem{AI-IMU}
M.~Brossard, A.~Barrau, and S.~Bonnabel, ``Ai-imu dead-reckoning,'' \emph{IEEE Transactions on Intelligent Vehicles}, vol.~5, no.~4, pp. 585--595, 2020.

\bibitem{RL-AKF}
\BIBentryALTinterwordspacing
X.~Gao, H.~Luo, B.~Ning, F.~Zhao, L.~Bao, Y.~Gong, Y.~Xiao, and J.~Jiang, ``Rl-akf: An adaptive kalman filter navigation algorithm based on reinforcement learning for ground vehicles,'' \emph{Remote. Sens.}, vol.~12, p. 1704, 2020. [Online]. Available: \url{https://api.semanticscholar.org/CorpusID:219711004}
\BIBentrySTDinterwordspacing

\bibitem{Fuzzy}
\BIBentryALTinterwordspacing
F.~Yan, S.~Li, E.~Zhang, and Q.~Chen, ``An intelligent adaptive kalman filter for integrated navigation systems,'' \emph{IEEE Access}, vol.~8, pp. 213\,306--213\,317, 2020. [Online]. Available: \url{https://api.semanticscholar.org/CorpusID:228091710}
\BIBentrySTDinterwordspacing

\bibitem{A-KIT}
\BIBentryALTinterwordspacing
N.~Cohen and I.~Klein, ``A-kit: Adaptive kalman-informed transformer,'' \emph{ArXiv}, vol. abs/2401.09987, 2024. [Online]. Available: \url{https://api.semanticscholar.org/CorpusID:267034974}
\BIBentrySTDinterwordspacing

\bibitem{TLIO}
\BIBentryALTinterwordspacing
W.~Liu, D.~Caruso, E.~Ilg, J.~Dong, A.~I. Mourikis, K.~Daniilidis, V.~R. Kumar, and J.~J. Engel, ``Tlio: Tight learned inertial odometry,'' \emph{IEEE Robotics and Automation Letters}, vol.~5, pp. 5653--5660, 2020. [Online]. Available: \url{https://api.semanticscholar.org/CorpusID:220363628}
\BIBentrySTDinterwordspacing

\bibitem{MT-AKF}
\BIBentryALTinterwordspacing
F.~Wu, H.~Luo, H.~Jia, F.~Zhao, Y.~Xiao, and X.~Gao, ``Predicting the noise covariance with a multitask learning model for kalman filter-based gnss/ins integrated navigation,'' \emph{IEEE Transactions on Instrumentation and Measurement}, vol.~70, pp. 1--13, 2021. [Online]. Available: \url{https://api.semanticscholar.org/CorpusID:228089808}
\BIBentrySTDinterwordspacing

\bibitem{Trainable_Quaternion}
\BIBentryALTinterwordspacing
G.~Milam, B.~Xie, R.~Liu, X.~Zhu, J.~Park, G.-W. Kim, and C.~H. Park, ``Trainable quaternion extended kalman filter with multi-head attention for dead reckoning in autonomous ground vehicles,'' \emph{Sensors (Basel, Switzerland)}, vol.~22, 2022. [Online]. Available: \url{https://api.semanticscholar.org/CorpusID:252868813}
\BIBentrySTDinterwordspacing

\bibitem{Du2023Agrobot}
\BIBentryALTinterwordspacing
Y.~Du, S.~S. Saha, S.~S. Sandha, A.~Lovekin, J.~Wu, S.~Siddharth, M.~Chowdhary, M.~K. Jawed, and M.~Srivastava, ``Neural-kalman gnss/ins navigation for precision agriculture,'' \emph{2023 IEEE International Conference on Robotics and Automation (ICRA)}, pp. 9622--9629, 2023. [Online]. Available: \url{https://api.semanticscholar.org/CorpusID:259337117}
\BIBentrySTDinterwordspacing

\bibitem{esfahani2019aboldeepio}
M.~A. Esfahani, H.~Wang, K.~Wu, and S.~Yuan, ``Aboldeepio: A novel deep inertial odometry network for autonomous vehicles,'' \emph{IEEE Transactions on Intelligent Transportation Systems}, vol.~21, no.~5, pp. 1941--1950, 2019.

\bibitem{zhang2021prediction}
G.~Zhang, P.~Xu, H.~Xu, and L.-T. Hsu, ``Prediction on the urban gnss measurement uncertainty based on deep learning networks with long short-term memory,'' \emph{IEEE Sensors Journal}, vol.~21, no.~18, pp. 20\,563--20\,577, 2021.

\bibitem{revach2022kalmannet}
G.~Revach, N.~Shlezinger, X.~Ni, A.~L. Escoriza, R.~J. Van~Sloun, and Y.~C. Eldar, ``Kalmannet: Neural network aided kalman filtering for partially known dynamics,'' \emph{IEEE Transactions on Signal Processing}, vol.~70, pp. 1532--1547, 2022.

\bibitem{zhang2014loam}
J.~Zhang, S.~Singh, \emph{et~al.}, ``Loam: Lidar odometry and mapping in real-time.'' in \emph{Robotics: Science and systems}, vol.~2, no.~9.\hskip 1em plus 0.5em minus 0.4em\relax Berkeley, CA, 2014, pp. 1--9.

\end{thebibliography}

\end{document}